%% file: main.tex
\documentclass[runningheads]{llncs}
\usepackage[T1]{fontenc}
\usepackage{graphicx}
\usepackage{multirow}
\usepackage{colortbl}
\begin{document}
%
\title{Evaluating Answer Reranking Strategies in Time-sensitive Question Answering}

\author{Mehmet Kardan \and Bhawna Piryani \and Adam Jatowt \\
    \institute{University of Innsbruck, Austria} 
    \email{\{Mehmet.Kardan, Bhawna.Piryani, Adam.Jatowt\}@student.uibk.ac.at}}

%

%

\maketitle

%
%
              
%
\begin{abstract}
Despite advancements in state-of-the-art models and information retrieval techniques, current systems still struggle to handle temporal information and to correctly answer detailed questions about past events. In this paper, we investigate the impact of temporal characteristics of answers in Question Answering (QA) by exploring several simple answer selection techniques.
Our findings emphasize the role of temporal features in selecting the most relevant answers from diachronic document collections and highlight differences between explicit and implicit temporal questions.

\end{abstract}

\input{introduction.tex}
\input{approaches}

\input{results_n_conclusion.tex}

%
%
%
\bibliographystyle{splncs04}
\bibliography{bibtex}
%




\end{document}

%% file: introduction.tex
\section{Introduction}

QA systems specifically designed to operate on temporal document collections, spanning multiple years or even decades, remain relatively rare. Yet, these systems offer a significant advantage: the ability to answer questions about past events, even those concerning specific details not found in widely used resources. For instance, a question \textit{How old was Giorgio Gallara when he was shot?} may not be answered by the Wikipedia but can be resolved when searching in an archival news collection, such as The New York Times Archive \cite{NYTCorpus}.

One of the key challenges in such systems is understanding how to effectively incorporate the temporal characteristics of both documents and user queries to return the most accurate answers. Although the temporal dimension has been explored in traditional keyword-based search engines (i.e., Temporal Information Retrieval (IR) \cite{Alonso,TemporalInfoSurvey}), studies addressing temporality in QA remain limited. 
This is an area requiring deeper exploration.

\begin{figure}[]
\centering
\includegraphics[width=.75\textwidth]{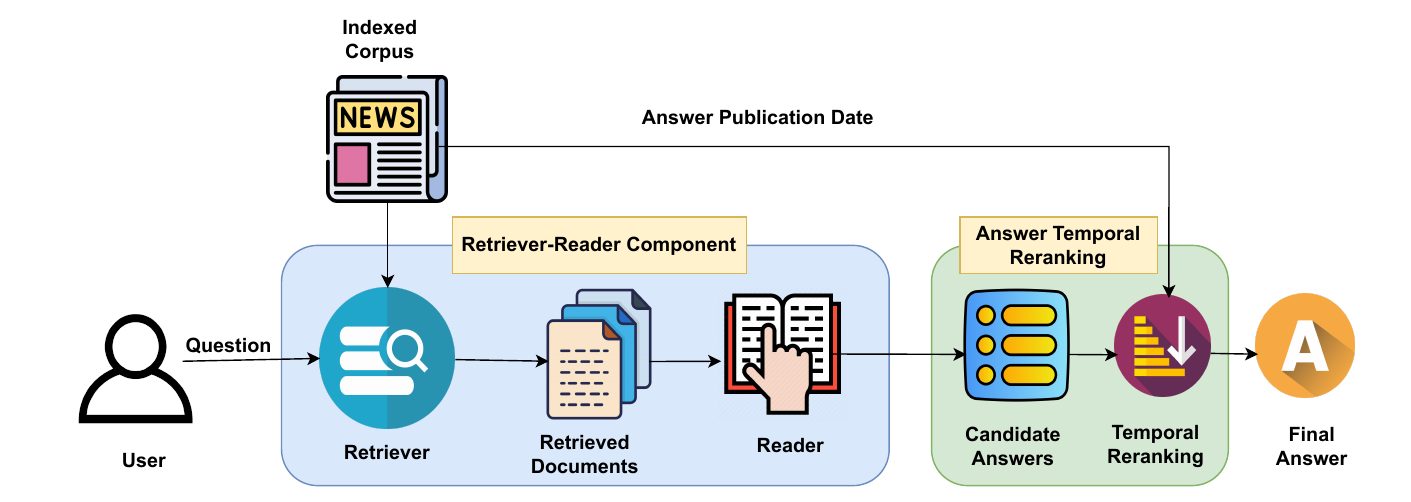}
\caption{Overview of the answer selection process applied in our analysis.} \label{fig:framework}
\end{figure}

In Temporal IR, document publication dates and the temporal characteristics of queries such as explicit dates embedded in their content
are typically matched during document ranking \cite{Alonso,TemporalInfoSurvey,jatowt2015mapping,cheng2020predicting,kawai2010chronoseeker}. 
However it is less clear 
what kind of temporal selection strategies should be used for answers 
when the input queries lack any explicit temporal markers
and the underlying collection spans a significant length of time. 
For example, should temporal QA systems prioritize then the most recent answers, or ones occurring in bursts within short time intervals,
or perhaps answers that are most common overall? 

In this work, we examine how the temporal characteristics of candidate answer impact the performance of QA systems. In other words, we are interested in the effect of temporal reranking strategies of candidate answers on correctness of question answering over diachronic corpora.
To this end, we evaluate multiple temporal reranking approaches using two QA datasets featuring time-sensitive questions. We use simple retrieval and answer extraction approaches like BM25 and BERT-based reader instead of LLMs and Retrieval-augmented Generation (RAG) to minimize the risk of hallucinations and errors and make the analysis transparent and interpretable. 

\section{Motivating Example}

\begin{figure}[!t]
\centering
\includegraphics[width=.7\textwidth]{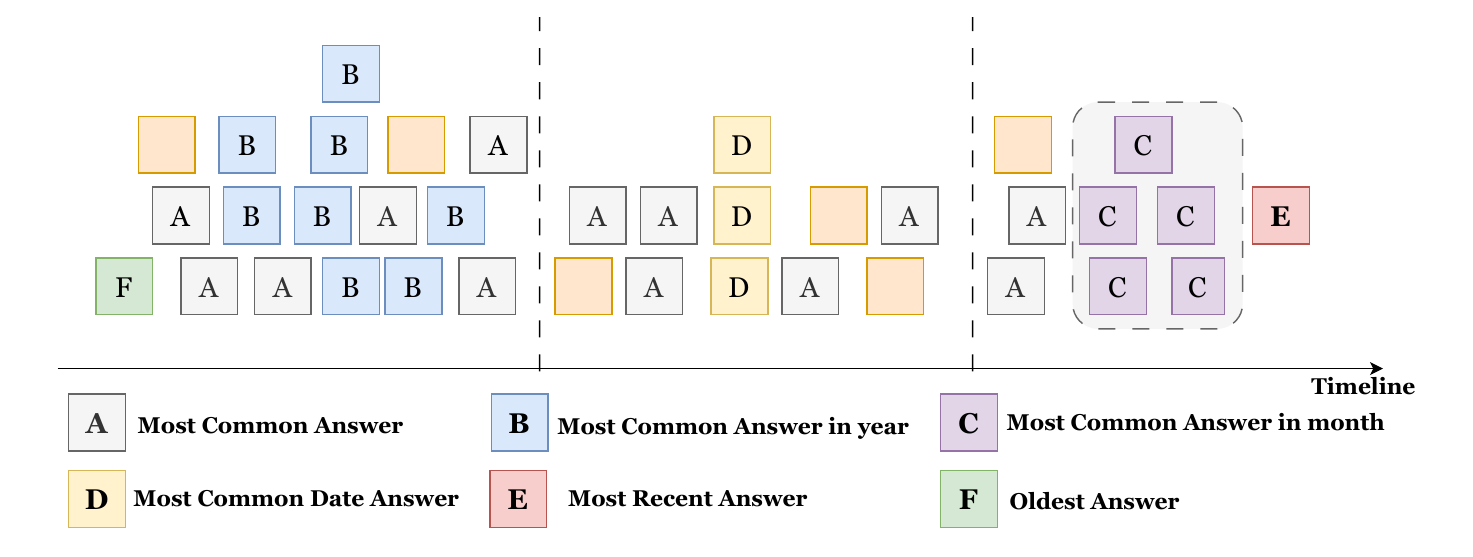}
\caption{An example of the temporal positioning of candidate answers, where the dotted line marks the division between years on the timeline, and the dotted box indicates here on particular month.
} 
\label{fig:Answerdistribution}
\end{figure}

To illustrate the analyzed problem, let us consider a hypothetical scenario, depicted in Figure \ref{fig:Answerdistribution}, where a user question about some past event has six possible candidate answers, each with its particular temporal distribution. 
The characteristics of potential answers are as follows:

\begin{itemize}
    \item Answer \textbf{\emph{A}} is the most commonly occurring across the entire timeline.
    
    \item Answer \textbf{\emph{B}} is the most commonly appearing answer in a single year.
    \item Answer \textbf{\emph{C}} is the most common answer in a single month.
    \item Answer \textbf{\emph{D}} is the most common answer at a particular date.
    
    \item Answer \textbf{\emph{E}} is the most recent answer.
    \item Answer \textbf{\emph{F}} is the oldest answer.
\end{itemize}


Each candidate answer offers a particular reason for its selection. 
For example,  Answer \textbf{\emph{A}} appears promising due to its high overall frequency, a common strategy in answer post-processing and final selection in QA systems \cite{zhu2021retrieving}. Answer \textbf{\emph{B}}, being the most common answer within a single year, and Answer \textbf{\emph{C}}, the most common within a single month, exhibit bursty characteristics suggesting potential developments related to the queried event. Similarly, Answer \textbf{\emph{D}}, which occurs multiple times within just a single day, might indicate the possible time of the event's occurrence or the time soon after it.
Answers B, C and D may thus provide specific details about the event, which may not be reported afterwards.
Answer \textbf{\emph{E}}, as the most recent, might provide the latest insights benefiting from the hindsight, though its low frequency could raise reliability concern. Finally, Answer \textbf{\emph{F}}, the chronologically first answer, although  less likely to be correct,
might still sometimes be the only one containing correct information about the event due to its primacy characteristic. 

We note that these are just particular cases and in reality, there will be different variations or combinations of those. Nevertheless, the question as to what extent the temporal positions of answers matter in relation to the occurrence of queried events is still open.
%
%
Investigating temporal answer distributions in diachronic document collections could present valuable opportunities for improving the accuracy and robustness of open-domain QA systems implemented over such corpora.

\section{Experimental Settings}
\vspace{-3mm}
We follow the answer selection process depicted in Figure 1.
As our document collection, we utilize the New York Times Annotated Corpus (NYT corpus) \cite{NYTCorpus} containing 1.8 million articles from 1987 to 2007. For QA datasets, we employ the subset of ArchivalQA \cite{ArchivalQA}, which in total consists of 15k question-answer pairs covering major and minor events, and TemporalQuestions \cite{QANA} dataset, which includes 1,000 time-specific questions. Both datasets cover the same period as the NYT corpus. For each dataset, we use two types of questions: those with explicit dates (Explicit QS) and those lacking them (Implicit QS). The datasets' statistics are detailed in Table \ref{table:1}.
\vspace{-4mm}

\begin{table}[]
\centering
\caption{Dataset statistics (the lenghts are represented in word units).}
\vspace{-4mm}
\setlength\tabcolsep{6pt}
\resizebox{1\textwidth}{!}{
\begin{tabular}{c|c|c|c|c}
\hline
\textbf{Dataset Name} & \textbf{Sub-Type} & \textbf{\begin{tabular}[c]{@{}c@{}}\# of \\ Questions\end{tabular}} & \textbf{\begin{tabular}[c]{@{}c@{}}Max/Min/Average\\ Question Length\end{tabular}} & \textbf{\begin{tabular}[c]{@{}c@{}}Max/Min/Average\\ Answer Length\end{tabular}} \\
\hline
\multirow{2}{*}{ArchivalQA}
& \begin{tabular}[c]{@{}c@{}}Explicit QS\end{tabular} & 7,500 & 30/5/11.36 & 16/1/1.88 \\ \cline{2-5} 
& \begin{tabular}[c]{@{}c@{}}Implicit QS\end{tabular} & 7,500 & 28/5/10.85 & 16/1/1.84 \\ \hline
\multirow{2}{*}{TemporalQuestions} & \begin{tabular}[c]{@{}c@{}}Explicit QS\end{tabular} & 500 & 36/5/13.37 & 15/1/2.06 \\ \cline{2-5} 
& \begin{tabular}[c]{@{}c@{}}Implicit QS\end{tabular} & 500 & 33/6/13.96 & 8/1/2.03 \\ \hline
\end{tabular}}
\label{table:1}
\end{table}
\vspace{-5mm}


To identify the optimal retrieval method, we first compare BM25 (using BERTSerini \cite{BERTSerini} implementation) and Dense Passage Retrieval (DPR) \cite{DPR} on 1,000 sampled ArchivalQA questions. The NYT collection has been prepared by dividing its documents into uniform length passages 
resulting in 1,194,730 passages. We then assign to each such passage the publication date of its parent document. 
We found that BM25 outperformed DPR, achieving 64.60 EM and 75.37 F1 on ArchivalQA and 33.40 EM and 44.27 F1 on TemporalQuestions, which led us to select it for further experiments. 

In our analysis, we follow BERTSerini’s setup \cite{BERTSerini}, retrieving the top 100 passages to have about 86\% chance of capturing relevant content (as stated in \cite{BERTSerini}), and using BERT \cite{devlin-etal-2019-bert} as the reader fine-tuned on SQuAD \cite{rajpurkar-etal-2016-squad}. We set the mixing weight 
\emph{µ} to 0.5 for balancing the retriever and reader scores as in Eq. \ref{equation:1}. 
\vspace{-2mm}
\begin{equation} 
\label{equation:1} 
Combined Score = (1 - \mu) * S_{BM25} + \mu * S_{BERT} 
\end{equation}

Using BERT as a reader we extract answers from each of the top returned 100 passages.
In the subsequent sections, we will explore different ways of selecting the best answer for each question from among all its extracted candidate answers.

%% file: approaches.tex
\vspace{-4mm}
\section{\textbf{Non-Temporal Answer Reranking Approaches}}
\vspace{-2mm}
This section presents four basic answer reranking methods that exclude temporal information, relying solely on retrieval and reader outputs. They serve as reference for comparison with the temporal approaches introduced in Section 5.

\textbf{Retrieval Based Reranking:}
In this approach, we rank answers based exclusively on retrieval scores, specifically using BM25 scores. By setting $\mu$ to 0 in Eq. \ref{equation:1}, we rely only on the retriever’s output to determine the final answer ranking. 

\textbf{Reader Based Reranking:}
This reranking method relies solely on reader scores after we extract candidate answer spans, achieved by setting $\mu$ to 1 in Eq. \ref{equation:1}. We select the final answer based on the highest reader score.

\textbf{Most Common Answer:}
This method selects the answer span that appears most frequently across retrieved passages, assuming that if a particular answer appears many times, it is more likely to be correct.

\textbf{Hybrid Retrieval-Reader:}
In this approach, we rank answers by combining retriever and reader scores, with $\mu$ in Eq. \ref{equation:1} set to 0.5, giving equal weight to both. This is in fact the original ranking.
\vspace{-4mm}

\section{Temporal Answer Reranking Approaches}
\vspace{-2mm}
This section describes approaches incorporating time as a reranking factor for retrieved answers.

\vspace{-4mm}
\subsection{\textbf{Direct Temporal Reranking Methods}}
The initial temporal reranking methods rely on publication dates without using complex algorithms, aiming to establish reference performance of how temporal information alone can influence the answer selection. 

\textbf{Most Recent Answer:} In this approach, we rank answers based on the publication dates of their passages, selecting a top answer that is the most recent. The motivation is that such an answer might contain the latest and most up-to-date information. For example, following a major event like an earthquake, we expect later documents to provide more accurate details on casualties and damage. Thus, selecting answers from the most recent document may yield up-to-date information.
 
\textbf{Oldest Answer:} For completeness, we also consider selecting the answer from the passage with the oldest timestamp in the retrieved set. This method is meant to serve as a reference allowing to assess to what extent the earlier content could offer useful insights for temporal questions. 

\textbf{Most Common Date Answer:} Similar to the non-temporal approach of selecting the most common answer, here we select the most common date from the timestamps of the retrieved passages using a day granularity. We follow an assumption that such a date may indicate the time when the questioned event occurred, or be at least near to it. If multiple days have the same frequency, the earliest one is chosen.

\vspace{-3mm}
\subsection{\textbf{Answer Reranking by Temporal Grouping}}
\vspace{-2mm}
We also explore methods that group answers by their publication dates using larger time units like months or years. The motivation behind this ranking is to explore the intuition that answers from top-ranked documents published in close time (e.g., ones likely forming bursts) may represent significant developments related to the input question.
We experiment with a simple time-based binning of answers, which is computationally inexpensive.

\textbf{Temporal Monthly Grouping:} In this method, we identify the month with the highest candidate answers. Within that month, we rank answers based on 
Hybrid Retrieval-Reader scores and select the top-scored answer.

\textbf{Temporal Yearly Grouping:} Following a similar process, we group candidate answers by the years of their publication dates. We select the year with the largest number of answers and, same as above, we extract the highest-scored answer within that group.

%% file: results_n_conclusion.tex
\vspace{-3mm}
\section{Results}
\vspace{-3mm}
Table \ref{table:2} shows the performance results of the non-temporal methods. On the ArchivalQA dataset, the \texttt{Retrieval-based Reranking} method performs well with Exact Match (EM) and F1 scores of 42.97 and 57.96, respectively, for explicit questions, and 41.51 EM and 56.63 F1 for implicit questions. However, on the TemporalQuestions, the \texttt{Reader-based Reranking} outperforms the retrieval-based approach, achieving an EM of 26.60 and F1 of 37.38 for explicit questions, and an EM of 36.60 and F1 of 46.25 for implicit questions. Across both datasets, the \texttt{Hybrid Retrieval-Reader} method combining the reader and retriever scores yields the best overall performance, achieving on ArchivalQA's explicit questions EM and F1 scores of 50.05 and 66.47, respectively, and 39.60 and 51.48 for implicit questions of TemporalQuestions dataset.

Table \ref{table:5} summarizes the performance of the direct temporal methods. Among these, the \texttt{Most Common Date Answer} approach consistently achieves the best performance, with an EM of 19.49 and F1 of 27.72 on ArchivalQA explicit questions and 26.20 and 36.43 on implicit questions of TemporalQuestions. Comparatively, the \texttt{Most Recent Answer} approach performs better than the \texttt{Oldest Answer} approach, aligning with the expectation that recent documents provide more accurate and relevant information for event-based questions.

Table \ref{table:6} presents results from temporal grouping approaches, with \texttt{Temporal Yearly Grouping} outperforming \texttt{Temporal Monthly Grouping}. For instance, on ArchivalQA's explicit questions, the \texttt{Temporal Yearly Grouping} method achieves an EM of 33.27 and F1 of 45.94, and on the TemporalQuestions dataset, it attains an EM of 31.20 and F1 of 40.95 for explicit questions. This suggests that grouping answers by wider timeframes can better capture relevant temporal information.

The results suggest that non-temporal reranking approaches, like retrieval-based and reader-based methods, generally outperform temporal methods for both explicit and implicit questions. Future work should however explore the combination of temporal and non-temporal answer rerankings. 
A combined approach could leverage the strengths of both, improving QA accuracy in complex, diachronic document collections. Notably, explicit and implicit temporal questions yield different performance levels, suggesting that temporal QA methods should address these two types separately.


\vspace{-4mm}
\begin{table}[]
\centering\setlength\tabcolsep{6pt}
\caption{Performance comparison of non-temporal approaches. }
\vspace{-4mm}
\resizebox{1\textwidth}{!}{
\begin{tabular}{c|cccc|cccc}
\hline
& \multicolumn{4}{c|}{\textbf{ArchivalQA}} & \multicolumn{4}{c}{\textbf{TemporalQuestions}} \\ \cline{2-9}
& \multicolumn{2}{c|}{\begin{tabular}[c]{@{}c@{}}\textbf{Explicit QS}\end{tabular}}
& \multicolumn{2}{c|}{\begin{tabular}[c]{@{}c@{}}\textbf{Implicit QS}\end{tabular}}
& \multicolumn{2}{c|}{\begin{tabular}[c]{@{}c@{}}\textbf{Explicit QS}\end{tabular}}
& \multicolumn{2}{c}{\begin{tabular}[c]{@{}c@{}}\textbf{Implicit QS}\end{tabular}} \\ \cline{2-9}
& \multicolumn{1}{c|}{EM} & \multicolumn{1}{c|}{F1} & \multicolumn{1}{c|}{EM} & F1 & \multicolumn{1}{c|}{EM} & \multicolumn{1}{c|}{F1} & \multicolumn{1}{c|}{EM} & F1 \\ \hline
{\begin{tabular}[c]{@{}c@{}}Retriever-Based Reranking\end{tabular}} &\multicolumn{1}{c|} {42.97} &\multicolumn{1}{c|} {57.96} &\multicolumn{1}{c|} {41.51} &\multicolumn{1}{c|} {56.63}  &\multicolumn{1}{c|} {21.00} &\multicolumn{1}{c|} {29.25} &\multicolumn{1}{c|} {26.40} &\multicolumn{1}{c} {38.11} \\ \hline

{\begin{tabular}[c|]{@{}c@{}}Reader-Based Reranking\end{tabular}} &\multicolumn{1}{c|} {34.79} &\multicolumn{1}{c|}{47.92} &\multicolumn{1}{c|} {35.60} &\multicolumn{1}{c|} {48.47} &\multicolumn{1}{c|} {26.60} &\multicolumn{1}{c|} {37.38} &\multicolumn{1}{c|} {36.60} &\multicolumn{1}{c} {46.25} \\ \hline

{\begin{tabular}[c]{@{}c@{}}Most Common Answer\end{tabular}} &\multicolumn{1}{c|} {20.47} &\multicolumn{1}{c|} {26.67} &\multicolumn{1}{c|} {17.60} &\multicolumn{1}{c|} {23.44} &\multicolumn{1}{c|} {32.20} &\multicolumn{1}{c|} {41.41} &\multicolumn{1}{c|} {35.20} &\multicolumn{1}{c} {47.08}  \\ \hline

{\begin{tabular}[c]{@{}c@{}}Hybrid Retrieval-Reader \end{tabular}} &\multicolumn{1}{c|} {\textbf{50.05}} &\multicolumn{1}{c|} {\textbf{66.47}} &\multicolumn{1}{c|} {\textbf{50.07}} &\multicolumn{1}{c|} {\textbf{67.10}} &\multicolumn{1}{c|} {\textbf{33.40}} &\multicolumn{1}{c|} {\textbf{44.27}} &\multicolumn{1}{c|} {\textbf{39.60}} & \multicolumn{1}{c}{\textbf{51.48}}  \\ 
\hline
\end{tabular}}
\label{table:2}
\end{table}
\vspace{-11mm}

\begin{table}[]
\centering
\caption{Comparison of direct temporal approaches.}
\vspace{-4mm}
\setlength\tabcolsep{6pt}
\resizebox{1\textwidth}{!}{
\begin{tabular}{c|cccc|cccc}
\hline
& \multicolumn{4}{c|}{\textbf{ArchivalQA}} & \multicolumn{4}{c}{\textbf{TemporalQuestions}} \\ \cline{2-9}
& \multicolumn{2}{c|}{\begin{tabular}[c]{@{}c@{}}\textbf{Explicit QS}\end{tabular}}
& \multicolumn{2}{c|}{\begin{tabular}[c]{@{}c@{}}\textbf{Implicit QS}\end{tabular}}
& \multicolumn{2}{c|}{\begin{tabular}[c]{@{}c@{}}\textbf{Explicit QS}\end{tabular}}
& \multicolumn{2}{c}{\begin{tabular}[c]{@{}c@{}}\textbf{Implicit QS}\end{tabular}} \\ \cline{2-9}
& \multicolumn{1}{c|}{EM} & \multicolumn{1}{c|}{F1} & \multicolumn{1}{c|}{EM} & F1 & \multicolumn{1}{c|}{EM} & \multicolumn{1}{c|}{F1} & \multicolumn{1}{c|}{EM} & F1 \\ \hline
\begin{tabular}[c]{@{}c@{}}Most Recent Answer\end{tabular} 
& \multicolumn{1}{c|}{4.39} & \multicolumn{1}{c|}{7.34} 
& \multicolumn{1}{c|}{4.43} & 7.03 
& \multicolumn{1}{c|}{6.80} & \multicolumn{1}{c|}{10.40} 
& \multicolumn{1}{c|}{10.00} & 14.76 \\ \hline
\begin{tabular}[c]{@{}c@{}}Oldest Answer\end{tabular} 
& \multicolumn{1}{c|}{3.68} & \multicolumn{1}{c|}{6.32} 
& \multicolumn{1}{c|}{3.68} & 5.98 
& \multicolumn{1}{c|}{3.40} & \multicolumn{1}{c|}{5.82} 
& \multicolumn{1}{c|}{2.60} & 5.01 \\ \hline
\begin{tabular}[c]{@{}c@{}}Most Common Date Answer\end{tabular}
& \multicolumn{1}{c|}{\textbf{19.49}} & \multicolumn{1}{c|}{\textbf{27.72}}
& \multicolumn{1}{c|}{\textbf{19.04}} & \textbf{26.89} 
& \multicolumn{1}{c|}{\textbf{18.40}} & \multicolumn{1}{c|}{\textbf{26.85}} 
& \multicolumn{1}{c|}{\textbf{26.20}} & \textbf{36.43} \\ \hline
\end{tabular}}
\label{table:5}
\end{table}
\vspace{-11mm}

\begin{table}[]
\centering
\caption{Results on Temporal Grouping based on monthly and yearly granularity.}
\vspace{-4mm}
\setlength\tabcolsep{6pt}
\resizebox{1\textwidth}{!}{
\begin{tabular}{c|cccc|cccc}
\hline
& \multicolumn{4}{c|}{\textbf{ArchivalQA}} & \multicolumn{4}{c}{\textbf{TemporalQuestions}} \\ \cline{2-9}
& \multicolumn{2}{c|}{\begin{tabular}[c]{@{}c@{}}\textbf{Explicit QS}\end{tabular}}
& \multicolumn{2}{c|}{\begin{tabular}[c]{@{}c@{}}\textbf{Implicit QS}\end{tabular}}
& \multicolumn{2}{c|}{\begin{tabular}[c]{@{}c@{}}\textbf{Explicit QS}\end{tabular}}
& \multicolumn{2}{c}{\begin{tabular}[c]{@{}c@{}}\textbf{Implicit QS}\end{tabular}} \\ \cline{2-9}
& \multicolumn{1}{c|}{EM} & \multicolumn{1}{c|}{F1} & \multicolumn{1}{c|}{EM} & F1 & \multicolumn{1}{c|}{EM} & \multicolumn{1}{c|}{F1} & \multicolumn{1}{c|}{EM} & F1 \\ \hline
 \begin{tabular}[c]{@{}c@{}}Temporal Monthly Grouping\end{tabular} & \multicolumn{1}{c|}{24.87} & \multicolumn{1}{c|}{35.23} & \multicolumn{1}{c|}{24.57} & 34.23 & \multicolumn{1}{c|}{28.00} & \multicolumn{1}{c|}{37.08} & \multicolumn{1}{c|}{\textbf{33.60}} & \textbf{45.34} \\ \hline
 \begin{tabular}[c]{@{}c@{}}Temporal Yearly Grouping\end{tabular} & \multicolumn{1}{c|}{\textbf{33.27}} & \multicolumn{1}{c|}{\textbf{45.94}} & \multicolumn{1}{c|}{\textbf{30.79}} & \textbf{42.84} & \multicolumn{1}{c|}{\textbf{31.20}} & \multicolumn{1}{c|}{\textbf{40.95}} & \multicolumn{1}{c|}{32.60} & 44.25 \\ \hline
\end{tabular}}
\label{table:6}
\end{table}
\vspace{-10mm}

\section{Conclusion}

\vspace{-2mm}

In this work, we study the importance of temporal characteristics in QA systems and analyze the impact of temporal reranking strategies on answer selection. 
Our findings indicate that temporal grouping methods proved most effective among simpler temporal reranking approaches, highlighting that grouping answers by their publication dates can be useful.
Nevertheless, non-temporal approaches seem to perform most effectively in the particular study settings we use. This needs to be verified this on other temporal QA datasets. Finally, we demonstrate that explicit and implicit temporal questions benefit from different temporal ranking strategies suggesting that their different treatment may be beneficial. 

Future work should continue similar analysis on other temporal datasets, including also domain-specific datasets (e.g., Medical \cite{kim-etal-2024-medexqa} or Legal QA \cite{abdallah2023exploring}), and explore effective ways of combining temporal and non-temporal reranking approaches (e.g., \cite{abdallah2025extendingdensepassageretrieval}).